\documentclass[sigconf]{acmart}

\usepackage{multirow}
\usepackage{algorithm}
\usepackage{algorithmic}
\usepackage{tabularx}
\usepackage{amsmath}
\usepackage{mathtools}
\usepackage{amsthm}
\usepackage{multirow}
\usepackage{bm}
\usepackage{xcolor} 
\usepackage[table]{xcolor}
\definecolor{lightblue}{rgb}{0.85,0.93,1.0}
\definecolor{lightpurple}{rgb}{0.93,0.85,1.0} 

\usepackage{microtype}
\usepackage{graphicx}
\usepackage{subcaption}
\usepackage{float}         
\usepackage{booktabs} 

\AtBeginDocument{%
  }

\copyrightyear{2026}
\acmYear{2026}
\setcopyright{cc}
\setcctype{by}
\acmConference[MM '26]{Proceedings of the 34th ACM International Conference on Multimedia}{November 10--14, 2026}{Rio de Janeiro, Brazil}
\acmBooktitle{Proceedings of the 34th ACM International Conference on Multimedia (MM '26), November 10--14, 2026, Rio de Janeiro, Brazil}
\acmDOI{10.1145/3767308.3836486}
\acmISBN{979-8-4007-2213-4/2026/11}

\begin{document}

\title{Dynamic Distribution-Aware Uncertainty Tracking in Vision-Language Representation Learning}

\author{Ao Zhou}
\affiliation{%
\institution{State Key Laboratory of Novel Software Technology, Nanjing University}
\city{Nanjing}
\country{China}
}
\email{602025330043@smail.nju.edu.cn}

\author{Zhiwei Jiang}
\correspondingauthor
\authornote {Corresponding author}
\affiliation{%
\institution{State Key Laboratory of Novel Software Technology, Nanjing University}
\city{Nanjing}
\country{China}
}
\email{jzw@nju.edu.cn}

\author{Zifeng Cheng}
\affiliation{%
\institution{State Key Laboratory of Novel Software Technology, Nanjing University}
\city{Nanjing}
\country{China}
}
\email{chengzf@nju.edu.cn}
\author{Cong Wang}
\affiliation{%
\institution{State Key Laboratory of Novel Software Technology, Nanjing University}
\city{Nanjing}
\country{China}
}
\email{wang.c@nju.edu.cn}
\author{Shufan Yang}
\affiliation{%
\institution{State Key Laboratory of Novel Software Technology, Nanjing University}
\city{Nanjing}
\country{China}
}
\email{sfyang@smail.nju.edu.cn}
\author{Haoru Chen}
\affiliation{%
\institution{Independent Researcher}
\city{Hangzhou}
\country{China}
}
\email{haoru0521@gmail.com}
\author{Qing Gu}
\affiliation{%
\institution{State Key Laboratory of Novel Software Technology, Nanjing University}
\city{Nanjing}
\country{China}
}
\email{guq@nju.edu.cn}

\renewcommand{\shortauthors}{Ao Zhou et al.}

%
\begin{abstract}
Uncertainty Quantification (UQ) aims to measure the reliability of model predictions, serving as a critical safeguard for deploying Vision-Language Models (VLMs) in safety-critical scenarios. 
Post-hoc approaches are widely adopted due to their lightweight nature, mapping the outputs of VLMs to uncertainty measures through learnable modules or inductive summarization.
However, post-hoc approaches remain inherently confined to fitting the failure patterns of the source domain, ignoring the dynamic nature of test distributions. 
To address this challenge, we propose a Dynamic Distribution-Aware Uncertainty Quantification framework (DDA-UQ) that shifts the paradigm from static mapping to a dynamic distribution-aware process. During training, we leverage a Gaussian Mixture Model to model the VLMs' embedding space and extract distributional evidence, thereby dynamically deriving uncertainty estimates. During inference, the design dynamically responds to changes in the data distribution. Extensive experiments demonstrate that our approach significantly outperforms state-of-the-art methods.
\end{abstract}

%
%


%


%
\maketitle

\section{Introduction}

Large-scale pretrained vision-language models (VLMs) such as CLIP~\citep{clip} have emerged as a cornerstone of zero-shot learning, demonstrating remarkable performance across diverse applications ranging from image recognition~\citep{vlm_survey,kgcoop,cocoop,ao1} and segmentation~\citep{open-voc1,open-voc2,ao2} to retrieval~\citep{retrieval1,retrieval2} without requiring task-specific fine-tuning~\citep{vlm3,domain_calibrating,empirical_calibrating,blip}.
However, this remarkable performance often comes with the risk of overconfident mispredictions \citep{DOR}. 
Consequently, deploying VLMs in safety-critical scenarios, such as autonomous driving and medical diagnosis~\cite{autonomous_vehicle, autonomous_vehicle2,failure_prediction2,active_learning1,medical_diagnosis}, necessitates reliable Uncertainty Quantification (UQ)~\cite{uq1,uq2,uq_survey,uq_survey2} as a safeguard against potential failures.
Fundamentally, UQ serves as a proxy for failure prediction, outputting a normalized score in [0,1] where higher values indicate a greater risk of prediction error.

\begin{figure}[t]
    \centering
    \includegraphics[width=0.9\linewidth]{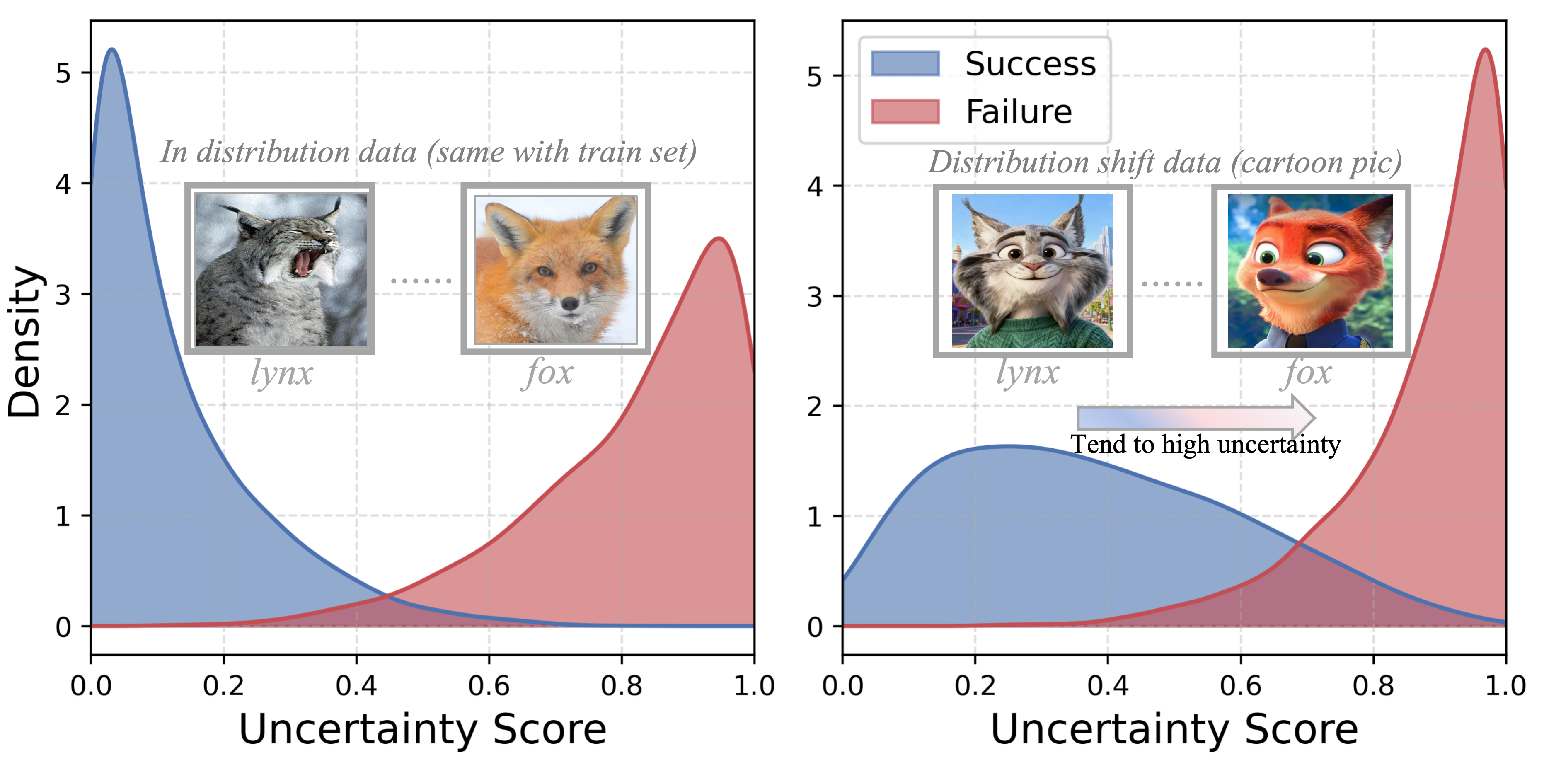}
    \caption{The SOTA post-hoc approach (e.g., ViLU) suffers from a significant limitation in generalization capability.} 
    \label{fig:intro}
\end{figure}
Unlike deep ensemble methods, such as MaskEnsembles~\cite{masksemble} and LoRA-Ensemble~\cite {uq3} that require costly retraining, post-hoc methods estimate uncertainty directly from pre-trained representations, offering a lightweight and deployment-friendly alternative. 
Within this category, traditional methods like Doctor~\citep{Doctor1} construct training-free statistical discriminators from softmax outputs, while Rel-U~\citep{relu} learns an uncertainty matrix to characterize confusion structures. 
However, Doctor is limited by the model's intrinsic calibration, and Rel-U tends to overfit the source domain's confusion topology. 
Consequently, under distribution shifts, these methods struggle to effectively capture uncertainties.
The state-of-the-art method, ViLU~\citep{vilu}, adopts a post-hoc training module. It utilizes the correctness of model predictions as a binary supervision signal to train a lightweight network that maps sample embeddings to risk scores based on training data. However, this approach relies on a \emph{static, sample-level} perspective. By formulating uncertainty quantification as a fixed mapping, it suffers from limited generalization capability. Consequently, its learned decision boundary often fails to transfer effectively, causing the method to tend to produce high uncertainty estimates regardless of whether the VLM's prediction is correct or incorrect, thereby limiting its effectiveness in real-world scenarios. As illustrated in Figure \ref{fig:intro}, the method generalizes well to in-distribution natural \textit{lynx} images but fails on cartoon \textit{lynx} images, resulting in invariably high uncertainty scores.

To transcend the limitations, we introduce the Dynamic Distribution Aware post-hoc Uncertainty Quantification framework (DDA-UQ), which reframes uncertainty estimation by shifting the perspective from a sample-centric view to a distribution-aware paradigm.
Specifically, distinct from previous approaches, we jointly model the sample’s embedding and its relative position within class‑conditional distributions.
We first model the VLM embedding space via a Gaussian Mixture Model (GMM) to capture the latent distribution. We extract an evidence vector from GMM comprising two metrics: (i) \emph{cognitive typicality} derived from negative log-likelihood; and (ii) \emph{aleatoric ambiguity} captured by posterior entropy. 
Then, we inject the evidence vector into visual representations to incorporate distributional context.
Finally, the visual representations integrated with the evidence vector are passed through a simple linear layer followed by an activation function to produce the uncertainty score. We employ semantic-aware soft targets as supervisory signals, and the overall process is optimized using a cross-entropy loss.
Moreover, our framework learns distribution-aware capabilities with strong generalization by synergistically maintaining a dynamic GMM and an evidence adaptation mechanism during training. 
This design enables real-time, data-driven adaptation at inference, thereby refining uncertainty quantification by dynamically adjusting to unlabeled test data.
The main contributions of our method are as follows:
\begin{itemize}
    \item We propose DDA-UQ, a lightweight framework that reformulates uncertainty estimation from a static, sample-level decision problem into a distribution-level modeling task within the embedding space, thereby grounding the quantification in intrinsic statistical geometry.
    \item During training, we dynamically update sample positional evidence and condition sample embeddings on this evidence. Simultaneously, we substitute rigid hard-label supervision (0/1) with soft targets decomposed into semantic severity and predictive fragility.  
    \item During inference, the framework operates in an online adaptive mode, continuously adapting to the test data stream by dynamically updating the sufficient statistics of the Gaussian Mixture Model using unlabeled samples.
\end{itemize}

Extensive experiments demonstrate that our method consistently outperforms static approaches, exhibiting superior generalization.


\section{Related Work}

Current approaches for uncertainty quantification in vision-language models predominantly follow a post-hoc paradigm. These methods do not modify or retrain the base model; instead, they estimate uncertainty directly from its raw outputs or by training an additional lightweight module. Although \textit{Deep Ensembles} \cite{uq3,MIMO,batchensemble} perform well for uncertainty quantification, their computational cost scales linearly with the ensemble size, and they lack the flexibility required for online deployment, making them unsuitable for large-scale architectures. Consequently, deep ensemble methods are excluded from the scope of our discussion.

\noindent\textbf{Output-Based.}
Methods in this category directly derive uncertainty signals from the model's inference outputs or latent representations. 
Standard approaches include Entropy~\cite{Shannon}, which measures the dispersion of predictive probabilities, and MCM~\cite{mcm}, which evaluates visual-textual alignment. The latter is often synergized with calibration methods~\cite{uq3,uq4} to refine reliability.
More advanced approaches delve into statistical and geometric properties: Doctor~\cite{Doctor1} constructs a training-free discriminator grounded in statistical hypothesis testing, while Rel-U~\cite{relu} adopts a geometric perspective, modeling uncertainty as the distance to the decision boundary. 
However, since these methods rely heavily on the static, intrinsic properties of the pre-trained model, they often struggle to adapt effectively to distribution shifts at test time.

\noindent\textbf{Probabilistic Modeling.}
Beyond point estimates, methods like ProbVLM~\cite{probVLM} and BayesVLM~\cite{BayesVLM} introduce probabilistic modeling via Gaussian or Laplace approximations. However, these approaches rely on \emph{static} parameterization fixed to the source domain, thereby failing to adapt to distribution shifts at test time. 

\noindent\textbf{Failure Prediction via Learned Risk Heads.}
Another line of work, exemplified by ViLU~\cite{vilu}, trains a lightweight module to predict model failure using prediction correctness as a supervised signal. 
However, this approach learns a \emph{static} mapping from embeddings to risk scores based solely on the source domain. 
Consequently, it implicitly overfits the training distribution and struggles to generalize to distribution shifts or unseen concepts.



\section{Preliminaries}
\subsection{Contrastive Vision-Language Models}
During the pre-training stage, CLIP~\cite{clip} learns aligned visual and textual representations by maximizing the cosine similarity between image and text embeddings through a contrastive objective on large-scale image--text pairs. Unlike conventional classifiers trained on closed-set labels, CLIP exploits open-set semantic supervision from natural language, enabling it to capture a broad spectrum of visual concepts. As a result, CLIP supports zero-shot classification at test time without any additional training.

Specifically, given a test image \(x\) belonging to one of \(C\) candidate categories, let \(\bm{v}(x)\) denote the visual embedding produced by the image encoder. For each class \(c\), a text embedding \(\bm{t}(c)\) is obtained by encoding a prompt such as ``a photo of \texttt{\{class c\}}'', where the placeholder is replaced with the corresponding category name. The zero-shot predictive probability for class \(c\) is then computed as
\begin{equation}
P(c \mid x) = \frac{\exp\big(\cos\big(\bm{v}(x), \bm{t}(c)\big)/\tau\big)}
{\sum_{c'=1}^{C} \exp\big(\cos\big(\bm{v}(x), \bm{t}(c')\big)/\tau\big)},
\label{zs_vlm}
\end{equation}
where \(\tau\) is CLIP's learned temperature parameter and \(\cos(\cdot,\cdot)\) denotes cosine similarity.

\begin{figure*}[t]
    \centering
    \includegraphics[width=0.9\linewidth]{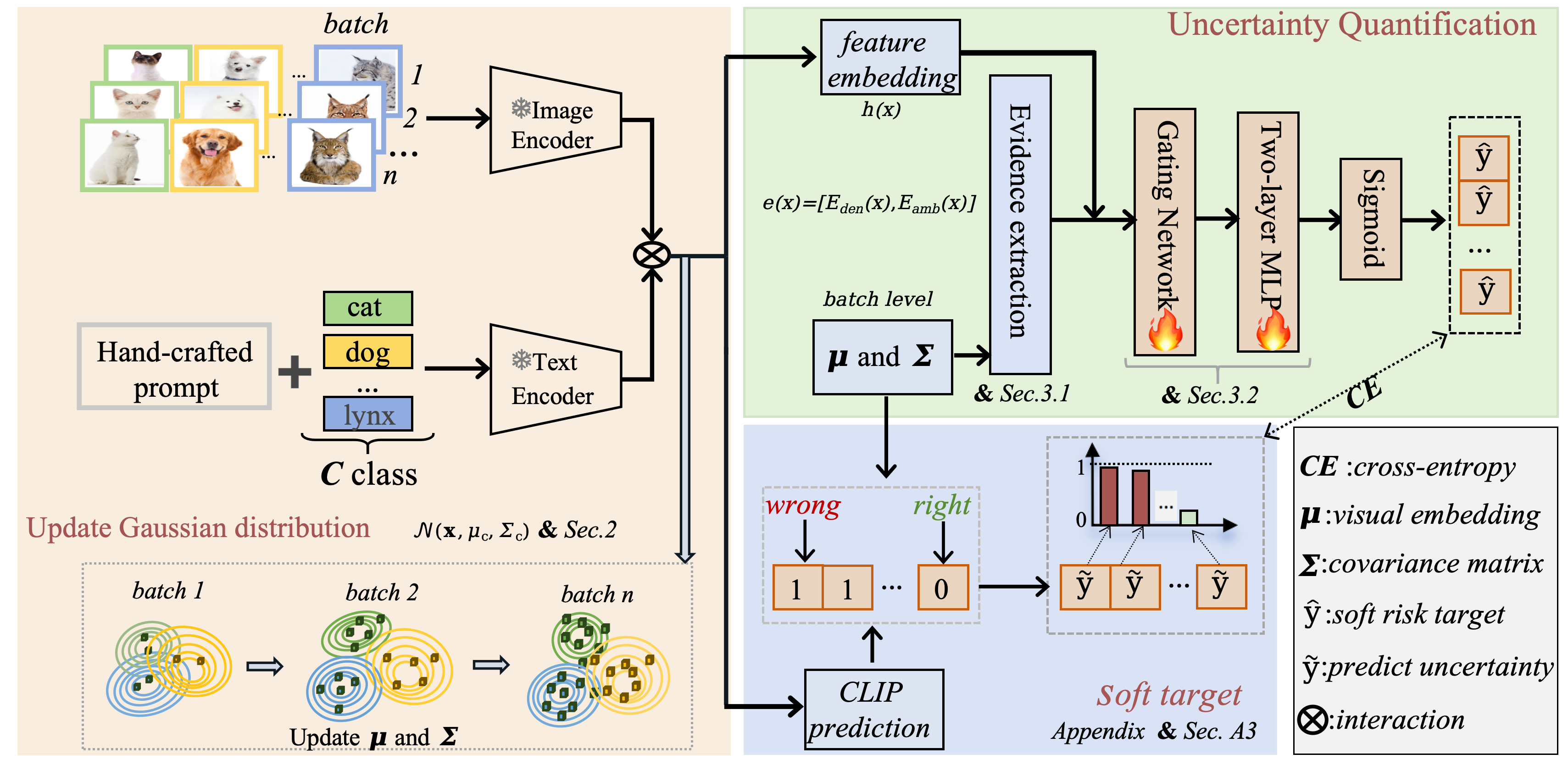}
    \caption{Framework overview. We model VLM outputs as a Gaussian mixture distribution updated with incoming streams. Extracted evidence is fused with visual embeddings and conditioned via a lightweight network to optimize uncertainty prediction using cross-entropy loss.} 
    \label{fig:framework}
\end{figure*}

\subsection{Gaussian Distribution Modeling in the Embedding Space}
\label{sec:GMM}
Recent studies have shown that CLIP representations exhibit pronounced
class-conditional clustering behavior in the embedding space, and can be
well approximated by multivariate Gaussian distributions
\cite{GDA1,GDA2,GDA3}. Motivated by these findings, we shift the perspective from evaluating a sample in isolation to assessing it within a collective statistical context. This motivates us to adopt a generative view inspired by Gaussian Discriminant Analysis (GDA), explicitly modeling the feature distribution to enable distribution-aware inference.

Specifically, we assume that the embedding distribution of each semantic class follows a multivariate Gaussian. Given an image $x \in \mathbb{R}^{d}$, the class-conditional \textbf{likelihood} for class $c$ is defined as:
\begin{equation}
P(x \mid c) = \mathcal{N}\big(x; \boldsymbol{\mu}_c, \boldsymbol{\Sigma}_c\big),
\label{eq:gmm_cond}
\end{equation}
where $\boldsymbol{\mu}_c$ and $\boldsymbol{\Sigma}_c$ denote the mean vector and covariance matrix of class $c$, respectively. Consequently, the \textbf{marginal distribution} of $x$ is formulated as a Gaussian mixture over all classes:
\begin{equation}
P(x) = \sum_{c=1}^{C} P(c) \cdot \mathcal{N}\big(x; \boldsymbol{\mu}_c, \boldsymbol{\Sigma}_c\big),
\label{eq:gmm_marginal}
\end{equation}
where $P(c)$ denotes the class prior. Under the uniform prior assumption $P(c)=1/C$, the \textbf{posterior} distribution is fully determined by the relative geometry between the embedding and the class-conditional Gaussian component:
\begin{equation}
P(c \mid x)
=
\frac{\mathcal{N}\!\big(x; \boldsymbol{\mu}_c, \boldsymbol{\Sigma}_c\big)}
     {\sum_{c'=1}^{C} \mathcal{N}\!\big(x; \boldsymbol{\mu}_{c'}, \boldsymbol{\Sigma}_{c'}\big)}.
\label{distribution_posterior}
\end{equation}

Consequently, the posterior can be reformulated into a normalized exponential (softmax) form
\begin{equation}
P(c\mid x) = \exp(f_c(x)) / \sum_{c'=1}^{C}\exp(f_{c'}(x)),
\end{equation}
where the discriminant function is
\begin{equation}
f_c(x)=
-\frac{1}{2}(x-\boldsymbol{\mu}_c)^{\top}\boldsymbol{\Sigma}_c^{-1}(x-\boldsymbol{\mu}_c)
-\frac{1}{2}\log|\boldsymbol{\Sigma}_c|+\log\pi_c.
\end{equation}
Intuitively, $f_c(x)$ quantifies the compatibility between the embedding $x$ and class $c$ by balancing two geometric factors: the quadratic term represents the Mahalanobis distance to the class mean, while the log-determinant term penalizes class-wise dispersion. This formulation ensures that the posterior is governed not just by proximity to a prototype, but by the intrinsic shape of the class distribution.
It is worth noting that, in our actual implementation, we adopt a \textbf{shared covariance matrix} $\boldsymbol{\Sigma} \in \mathbb{R}^{d \times d}$ across all semantic classes, instead of the class-specific covariance matrices $\boldsymbol{\Sigma}_c$ described above. This modification simplifies the model while maintaining the core distribution-aware inference capability. For explanations and reasoning regarding this part, see Appendix A.1 for details.

\section{Method}
\label{sec:method}

Existing UQ methods for VLMs typically rely on static estimates fixed by training data (e.g., Eq. \eqref{zs_vlm}). We argue this paradigm is fundamentally limited: by evaluating uncertainty in isolation based on rigid decision boundaries, these methods fail to generalize, leading to unreliable estimates under distribution shifts.
Leveraging the Gaussian framework (Sec.~\ref{sec:GMM}), we propose a shift from static sample-level evaluation to dynamic distributional adaptation. Our core insight is that UQ must be \emph{location-aware}—grounded in the sample's position within the feature manifold.

\subsection{Distributional Evidence}

Building on the Gaussian modeling in Sec.~\ref{sec:GMM}, we extract two complementary distributional evidences to characterize uncertainty: \textit{distributional typicality} and \textit{semantic ambiguity}.

We first quantify the support of a sample within the global embedding space using the negative log-likelihood (NLL) of the mixture density:
\begin{equation}
E_{\text{den}}(x) = - \log P(x).
\end{equation}
This metric assesses how well a sample conforms to the learned target manifold. Low NLL values indicate distributionally typical samples, whereas high values flag outliers in low-density regions, signaling potential out-of-distribution inputs. Crucially, $E_{\text{den}}$ captures global geometric support independent of specific class assignments.

While density assesses representativeness, it fails to capture separability; a sample can be distributionally typical yet ambiguous if it resides in the intersection of multiple Gaussian components. To isolate this, we compute the entropy of the classification posterior:
\begin{equation}
E_{\text{amb}}(x) =
- \sum_{c=1}^{C} P(c\mid x) \log P(c\mid x).
\end{equation}
Elevated entropy indicates that the sample lies near decision boundaries where class prototypes overlap, reflecting high aleatoric uncertainty. Conversely, minimized entropy corresponds to samples deeply embedded within a single class-specific cluster. 
Finally, we standardize these metrics to ensure scale consistency and aggregate them into a compact evidence vector $\mathbf{e}(x) = [\tilde{E}_{\text{den}}(x), \tilde{E}_{\text{amb}}(x)]^{\top}$.

\subsection{Framework}
After extracting evidence from the distribution, we combine the evidence with the representation vectors. 
Specifically, the visual embedding $\bm{v}(x)$ is first mapped to a hidden representation $\mathbf{h}(x) \in \mathbb{R}^d$ via a linear projection. Simultaneously, a gating network processes the evidence vector $\mathbf{e}(x)$ to generate sample-specific affine transformation parameters—namely, a scaling factor $\boldsymbol{\gamma}$ and a shift term $\boldsymbol{\beta}$:
\begin{equation}
\boldsymbol{\gamma} = \sigma(\mathbf{W}_{\gamma}\mathbf{e}(x) + \mathbf{b}_{\gamma}), \quad
\boldsymbol{\beta} = \mathbf{W}_{\beta}\mathbf{e}(x) + \mathbf{b}_{\beta},
\end{equation}
where $\mathbf{W}_{\gamma}, \mathbf{W}_{\beta} \in \mathbb{R}^{d \times 2}$ are learnable projection matrices and $\sigma$ denotes the sigmoid function. The visual features $\mathbf{h}(x)$  are then modulated as $\boldsymbol{\gamma} \odot \mathbf{h}(x) + \boldsymbol{\beta}$. The gating branch introduces only a negligible number of parameters, and the increased expressiveness comes primarily from the structural prior: the uncertainty quantification is forced to be a function of visual features conditioned on their distributional validity. This mechanism enables the distributional evidence to perform channel-wise recalibration of visual features. For instance, evidence of distributional anomaly (e.g., a high $E_{\text{den}}$ value) can suppress activations that lack support from the target distribution, thereby guiding the predictor to rely on more reliable signals.
Then, a two-layer MLP predicts the scalar uncertainty score from the modulated representation:
\begin{equation}
\hat{y} = \sigma\!\Bigl(\mathbf{w}_{2}^{\top}\phi \bigl(\mathbf{W}_{1}\mathbf{h}'(x)+\mathbf{b}_{1}\bigr)+b_{2}\Bigr),
\end{equation}
where \(\phi(\cdot)\) is the ReLU activation function, and \(\sigma(\cdot)\) ensures the output lies in \((0,1)\).

\subsection{Training Procedure and Update Process}
We optimize the predicted score $\hat{y}$ using a standard binary cross-entropy objective. Departing from conventional hard label, which assigns 0 to correct predictions and 1 to errors \cite{vilu}, we construct a continuous soft label $\tilde{y} \in [0,1]$ theoretically derived from the Gaussian discriminant distribution. 
This soft label integrates both the \emph{semantic severity} of misclassifications and the \emph{predictive fragility} of correct predictions. 
Specifically, for misclassified samples (hard label: 1), the risk is scaled by semantic severity; this design penalizes overconfident and semantically implausible errors more heavily while attenuating the risk for confusions between related concepts. 
Conversely, for correctly classified samples (hard label 0), we characterize fragility using the probability margin of top-2 predictions (detailed derivation in Appendix A.3). 
Given this soft label $\tilde{y}$ and the predicted score $\hat{y}$, we achieve calibrated uncertainty estimation by minimizing the following loss function:
\begin{equation} 
\mathcal{L} = - \frac{1}{s} \sum_{i=1}^{s} \big[ \tilde{y}_i \log(\hat{y}_i) + (1 - \tilde{y}_i) \log(1 - \hat{y}_i) \big]
\label{eq:bce_loss} 
\end{equation}
where $s$ denotes the batch size.

\begin{algorithm}[!h]
\caption{Dynamic Uncertainty Quantification}
\label{alg:batch_ddauq}
\begin{algorithmic}[1]
    \STATE \textbf{Input:} CLIP encoder $f(\cdot)$; Current Batch $\mathcal{S}$.
    
    \STATE $\{x_i\}_{i=1}^s \gets f(\mathcal{S})$ 
    \STATE Initialize GMM $\{\bm{\mu}_c, \bm{\Sigma}_c\}_{c=1}^C$
    \STATE \textcolor{blue}{Step 1. Dynamic GMM Update}
    \FOR{each class $c \in \{1, \dots, C\}$}
        \STATE Update $\bm{\mu}_c, \bm{\Sigma}_c$ via EMA using $\{x_i\}_{i=1}^s$ statistics
    \ENDFOR
    
    \STATE \textcolor{blue}{Step 2. Adaptive Evidence Extraction}
    \FOR{each embedding $x_i$}
        \STATE Compute $E_{\text{den}}$ and $E_{\text{amb}}$ via GMM
        \STATE $\mathbf{e}(x_i) \gets \text{Standardize}([E_{\text{den}}, E_{\text{amb}}])$
    \ENDFOR
    
    \STATE \textcolor{blue}{Step 3. Uncertainty Prediction}
    \STATE $\mathbf{h}'(x)$ $\gets$ Gating network $(\{\mathbf{h}(x_i)\} ,\{\mathbf{e}(x_i)\})_{i=1}^s$
    \STATE $\{\hat{y}_i\}_{i=1}^s \gets \text{Predictor} \{\mathbf{h}'(x)\}_{i=1}^s$  
    \STATE  Compute loss  and optimizer step
    \STATE \textbf{Output:} A well-trained framework
\end{algorithmic}
\end{algorithm}
Algorithm \ref{alg:batch_ddauq} outlines the training procedure of our dynamic uncertainty estimation framework from a batch perspective.
In Step 1, the model updates the Gaussian Mixture Model via a momentum-based Exponential Moving Average (EMA) \cite{EMA}. This procedure is mathematically equivalent to a streaming Expectation-Maximization (EM) \cite{EM1,EM2} process (see Appendix A.2).
By continuously synchronizing statistical estimates with the underlying feature manifold, this mechanism fosters robust evidence extraction from partial distributional observations, thereby enabling the framework to resiliently accommodate diverse distributional shifts.
To unify supervised training and unsupervised adaptation, we employ the discriminative mechanism from Eq.~\eqref{distribution_posterior} to define a gate responsibility $r_c(x)$:
\begin{equation}
r_c(x)
=
\frac{P(c\mid x)\, \exp\!\big(\lambda\,\mathbb{I}\{c=c_t\}\big)}
{\sum_{c'} P(c'\mid x)\, \exp\!\big(\lambda\,\mathbb{I}\{c'=c_t\}\big)},
\end{equation}
where $\mathbb{I}\{\cdot\}$ is the indicator function and $c_t$ denotes the target class, and $\lambda \ge 0$ controls the strength of supervision. These responsibilities serve as dynamic weights for updating the GMM statistics: a sample $x$ of batch $\mathcal{S}$ contributes to the sufficient statistics of class $c$ proportional to $r_c(x)$. During training ($\lambda>0$), this formulation softly biases the responsibilities toward the ground-truth label, driving the GMM parameters $\{\boldsymbol{\mu}_c, \boldsymbol{\Sigma}_c\}$ to converge to accurate class-conditional densities. 

Subsequently, Step 2 extracts and standardizes distributional evidence based on density and semantic ambiguity. 
In Step 3, a gating network conditions the visual features on this evidence, followed by a two-layer MLP predictor that outputs calibrated uncertainty scores aligned with the current distributional perspective.

During inference ($\lambda=0$), the supervision term vanishes and $r_c(x)$ reduces to the zero-shot posterior, enabling the model to \emph{self-update} its parameters based on its own predictions in a fully unsupervised manner. This design naturally endows the model with test-time adaptation capabilities. By exclusively updating the GMM statistics while keeping the prediction head frozen, the model efficiently calibrates distributional evidence to the target distribution, thereby avoiding computationally expensive gradient-based parameter fine-tuning.

\begin{table*}[ht]
\centering
\caption{UQ Performance on image-label datasets. The zero-shot Accuracy (in \%) of CLIP for each dataset is shown above the corresponding columns. AUC$\uparrow$ indicates higher is better, FPR95$\downarrow$ indicates lower is better.}
\label{tab:selected_results}
\resizebox{0.8\textwidth}{!}{%
\begin{tabular}{lcccccccccc}
\toprule
& \multicolumn{2}{c}{\textbf{Caltech101}} & \multicolumn{2}{c}{\textbf{FGVCAircraft}} & \multicolumn{2}{c}{\textbf{StanfordCars}} & \multicolumn{2}{c}{\textbf{SUN397}} & \multicolumn{2}{c}{\textbf{UCF101}} \\
& \multicolumn{2}{c}{91.4\%} & \multicolumn{2}{c}{18.1\%} & \multicolumn{2}{c}{60.1\%} & \multicolumn{2}{c}{62.1\%} & \multicolumn{2}{c}{61.6\%} \\
\cmidrule(lr){2-3} \cmidrule(lr){4-5} \cmidrule(lr){6-7} \cmidrule(lr){8-9} \cmidrule(lr){10-11}
\textbf{Method} & \textbf{AUC$\uparrow$} & \textbf{FPR95$\downarrow$} & \textbf{AUC$\uparrow$} & \textbf{FPR95$\downarrow$} & \textbf{AUC$\uparrow$} & \textbf{FPR95$\downarrow$} & \textbf{AUC$\uparrow$} & \textbf{FPR95$\downarrow$} & \textbf{AUC$\uparrow$} & \textbf{FPR95$\downarrow$} \\
\midrule
MCM & 87.3±0.25 & 69.5±0.98 & 76.4±0.48 & 82.1±1.05 & 80.9±0.32 & 72.8±1.12 & 79.2±0.41 & 76.3±0.87 & 83.5±0.28 & 69.4±0.92 \\
TS + MCM & 89.7±0.21 & 56.3±1.15 & 75.2±0.52 & 81.9±1.08 & 82.1±0.38 & 72.5±1.08 & 78.3±0.39 & 74.9±0.92 & 83.9±0.31 & 68.9±0.95 \\
Entropy & 85.4±0.34 & 79.5±0.87 & 73.6±0.58 & 84.2±0.95 & 78.9±0.41 & 76.8±1.21 & 75.2±0.47 & 79.3±1.05 & 82.9±0.36 & 73.6±0.88 \\
Doctor & 88.0±0.29 & 65.9±1.02 & 75.3±0.51 & 82.0±1.11 & 81.5±0.35 & 72.5±1.14 & 78.9±0.43 & 76.5±0.96 & 85.2±0.33 & 69.8±1.07 \\
Rel-U & 89.5±0.26 & 57.8±1.34 & 69.1±0.72 & 81.8±1.22 & 74.9±0.62 & 79.5±1.32 & 74.5±0.58 & 80.7±1.18 & 83.5±0.40 & 62.0±1.45 \\
ProbVLM & 90.5±0.24 & 51.6±1.52 & 74.1±0.53 & 82.9±1.03 & 79.0±0.49 & 74.9±1.15 & 77.3±0.51 & 77.5±0.99 & 87.9±0.22 & 54.2±1.68 \\
BayesVLM & 93.3±0.18 & 36.9±1.78 & 71.3±0.65 & 85.0±0.89 & 86.9±0.31 & 62.8±1.54 & 79.6±0.37 & 74.1±1.21 & 85.1±0.42 & 65.7±1.12 \\
ViLU & \textbf{95.9±0.12} & 17.6±1.95 & 81.7±0.41 & 72.4±1.58 & 89.6±0.25 & 47.3±1.82 & \textbf{88.1±0.28} & 49.6±1.67 & 95.2±0.15 & 21.3±1.89 \\
\rowcolor{lightblue}
Ours (Base) & 94.7±0.14 & 14.8±1.67 & 84.7±0.35 & 61.2±1.35 & 93.2±0.23 & 38.9±1.55 & 86.1±0.31 & 53.3±1.48 & 93.8±0.19 & 20.1±1.72 \\
\rowcolor{lightpurple}
Ours (Adaptation) & 95.1±0.13 & \textbf{15.1±1.58} & \textbf{85.9±0.32} & \textbf{59.0±1.42} & \textbf{94.3±0.21} & \textbf{39.2±1.46} & 87.2±0.26 & \textbf{49.1±1.52} & \textbf{96.6±0.11} & \textbf{16.9±1.65} \\
\bottomrule
\end{tabular}%
}
\end{table*}

\section{Experiments}
\subsection*{Experimental Setup}
We focus on uncertainty quantification for VLMs in zero-shot classification tasks, and consider five experimental scenarios: \emph{standard benchmark settings}, \emph{distribution shift}, \emph{comparison of test-time adaptation scenarios}, \emph{multi-label classification}, and \emph{OOD scenarios}. In addition, we provide ablation studies and qualitative visualizations to further demonstrate the efficacy of the proposed uncertainty quantification approach.

\textbf{Datasets.} We employ two types of benchmark datasets: (i) image--label datasets and (ii) image--caption datasets.

\textbf{Baselines.} To comprehensively evaluate our approach, we compare it against state-of-the-art methods discussed in the \textit{Related Work} section, including: (i) output-based post-hoc methods (MCM, TS+MCM, Entropy, Doctor, Rel-U), (ii) probabilistic modeling approaches (ProbVLM, BayesVLM), and (iii) training a lightweight module (ViLU).

\textbf{Implementation Details.} In the main experiments, we adopt CLIP ViT-B/32 as the default backbone network. We evaluate two variants of the framework: the Base version, which accesses the entire test set offline to estimate distribution statistics, and the Adaptation version, which corresponds to an online deployment scenario where samples arrive sequentially and the statistics are updated incrementally on a per-sample basis. We employ five-fold cross-validation and report the mean value along with the standard deviation. In our method, the covariance is shared, which avoids the computational and storage overhead on high-dimensional datasets such as ImageNet-1K.

\textbf{Metrics.} We evaluate our model using two widely adopted metrics in uncertainty quantification and failure detection: (i) the False Positive Rate at 95\% True Positive Rate (FPR95), and (ii) the Area Under the Receiver Operating Characteristic Curve (AUC). These metrics measure the model’s ability to identify high-uncertainty (or potentially erroneous) predictions.

\subsection*{Main Result}

\noindent\textbf{UQ on Standard Benchmarks.}
We evaluate failure prediction performance on five standard image-label benchmarks, where the training and test sets are stratified splits sampled from the same dataset. As shown in Table~\ref{tab:selected_results}, our framework achieves superior or comparable performance to state-of-the-art methods in most cases. Compared to post-hoc approaches that rely on fixed decision rules, such as MCM, Doctor, and BayesVLM, our learning-based method demonstrates significant performance gains. Our approach outperforms the current SOTA method ViLU on the majority of benchmarks. Although ViLU also employs a trainable failure predictor, it relies on a static mapping from embeddings to risk scores. In contrast, DDA-UQ explicitly models the statistical geometry of the embedding space, an advantage that becomes particularly pronounced in fine-grained classification tasks with high semantic ambiguity. On the FGVCAircraft dataset, where frequent confusion among visually similar aircraft variants leads to a low base accuracy of only 18.1\%, ViLU achieves an AUC of 81.7\%. By leveraging an aleatoric ambiguity measure derived from the entropy of a Gaussian mixture model, our method effectively identifies these ambiguous samples, improving the AUC to 85.9\% (a gain of +4.2 percentage points) and substantially reducing FPR95. 
Even on datasets where ViLU exhibits competitive performance, such as Caltech101, our method achieves a markedly lower FPR95 (e.g., 14.8 vs. 17.6 for Ours (Base), and 15.1 vs. 17.6 for Ours (Adaptation)), indicating that our risk estimation is more conservative and reliable for safety-critical applications.
Notably, our adaptive variant achieves superior or comparable performance to the base version across multiple datasets, suggesting that the streaming exponential moving average update mechanism not only enables efficient adaptation to test-time distribution shifts, but also acts as a robust regularizer by preventing the statistical parameters from being overly influenced by outliers in the full test set. Experimental results on image-caption datasets are reported in Appendix~A.4.

\begin{table*}[ht]
\centering
\caption{UQ Performance on image-label datasets with distribution shift.}
\label{tab:ds}
\resizebox{\textwidth}{!}{%
\begin{tabular}{lcccccccccccccc}
\toprule
 & \multicolumn{2}{c}{\textbf{ImageNet-1k}} & \multicolumn{2}{c}{\textbf{ImageNet-R}} & \multicolumn{2}{c}{\textbf{ImageNet-C}} & \multicolumn{2}{c}{\textbf{CIFAR-10}} & \multicolumn{2}{c}{\textbf{CIFAR-10-C}} & \multicolumn{2}{c}{\textbf{CIFAR-100}} & \multicolumn{2}{c}{\textbf{CIFAR-100-C}} \\
 & \multicolumn{2}{c}{66.7\%} & \multicolumn{2}{c}{74.3\%} & \multicolumn{2}{c}{46.8\%} & \multicolumn{2}{c}{88.3\%} & \multicolumn{2}{c}{75.2\%} & \multicolumn{2}{c}{68.6\%} & \multicolumn{2}{c}{55.8\%} \\
\cmidrule(lr){2-3} \cmidrule(lr){4-5} \cmidrule(lr){6-7} \cmidrule(lr){8-9} \cmidrule(lr){10-11} \cmidrule(lr){12-13} \cmidrule(lr){14-15}
\textbf{Method} & \textbf{AUC$\uparrow$} & \textbf{FPR95$\downarrow$} & \textbf{AUC$\uparrow$} & \textbf{FPR95$\downarrow$} & \textbf{AUC$\uparrow$} & \textbf{FPR95$\downarrow$} & \textbf{AUC$\uparrow$} & \textbf{FPR95$\downarrow$} & \textbf{AUC$\uparrow$} & \textbf{FPR95$\downarrow$} & \textbf{AUC$\uparrow$} & \textbf{FPR95$\downarrow$} & \textbf{AUC$\uparrow$} & \textbf{FPR95$\downarrow$} \\
\midrule
MCM & 80.8±0.35 & 71.3±1.05 & 78.3±0.42 & 65.2±1.12 & 72.1±0.48 & 81.6±1.18 & 89.3±0.41 & 52.8±0.87 & 83.8±0.63 & 62.3±0.94 & 82.1±0.52 & 67.9±1.03 & 77.5±0.71 & 74.1±1.12 \\
TS + MCM & 80.7±0.36 & 71.5±1.08 & 83.9±0.31 & 58.4±1.05 & 80.7±0.32 & 71.5±1.12 & 90.2±0.38 & 51.0±0.92 & 84.6±0.55 & 61.5±0.79 & 84.4±0.47 & 67.9±0.88 & 78.9±0.64 & 73.4±0.97 \\
Entropy & 78.3±0.42 & 76.8±1.15 & 75.8±0.48 & 69.4±1.22 & 70.5±0.52 & 84.2±1.28 & 88.2±0.45 & 60.4±1.15 & 80.9±0.72 & 69.2±1.08 & 79.3±0.61 & 72.5±1.21 & 74.1±0.82 & 78.6±1.33 \\
Doctor & 80.3±0.38 & 72.9±1.12 & 76.9±0.45 & 67.8±1.18 & 71.8±0.49 & 82.5±1.25 & 89.0±0.36 & 57.1±0.95 & 84.0±0.58 & 63.9±0.86 & 82.9±0.48 & 69.2±1.05 & 77.2±0.69 & 75.9±1.18 \\
Rel-U & 75.1±0.52 & 85.0±1.42 & 81.0±0.39 & 58.2±1.08 & 75.1±0.47 & 85.0±1.38 & 85.7±0.52 & 55.0±1.01 & 79.7±0.68 & 65.3±0.82 & 80.4±0.56 & 68.8±0.93 & 74.6±0.77 & 76.5±1.25 \\
ProbVLM & 78.7±0.41 & 77.0±1.18 & 81.2±0.37 & 62.3±1.15 & 75.6±0.46 & 79.8±1.22 & 96.1±0.28 & 21.8±0.65 & 89.5±0.35 & 35.1±0.73 & 79.8±0.43 & 69.1±0.98 & 75.2±0.54 & 74.0±0.91 \\
BayesVLM & 81.5±0.33 & 70.3±1.05 & 83.5±0.29 & 58.7±1.08 & 78.9±0.39 & 76.4±1.18 & 92.0±0.33 & 45.4±0.76 & 86.8±0.42 & 54.0±0.69 & 86.5±0.39 & 60.9±0.84 & 81.5±0.47 & 68.4±1.05 \\
ViLU & 89.5±0.22 & 47.4±1.58 & 78.2±0.44 & 66.8±1.25 & 73.4±0.51 & 86.3±1.45 & 97.9±0.22 & 8.2±0.53 & 95.0±0.26 & 16.9±0.45 & 91.0±0.31 & 36.0±0.78 & 86.2±0.38 & 43.1±0.62 \\
\rowcolor{lightblue}
Ours (Base) & 91.2±0.18 & 43.8±1.45 & 88.9±0.24 & 48.6±1.22 & 84.7±0.31 & 65.1±1.28 & 98.5±0.18 & 7.5±0.48 & 96.1±0.22 & 14.8±0.42 & 92.3±0.27 & 32.5±0.72 & 87.9±0.34 & 39.5±0.58 \\
\rowcolor{lightpurple}
Ours (Adaptation) & \textbf{92.4±0.15} & \textbf{38.6±1.32} & \textbf{90.5±0.19} & \textbf{43.2±1.15} & \textbf{86.3±0.27} & \textbf{58.9±1.22} & \textbf{99.0±0.15} & \textbf{6.8±0.42} & \textbf{97.2±0.19} & \textbf{12.5±0.38} & \textbf{93.8±0.23} & \textbf{29.1±0.65} & \textbf{89.4±0.29} & \textbf{35.2±0.52} \\
\bottomrule
\end{tabular}%
}
\end{table*}

\noindent\textbf{Robustness to Distribution Shift.}
As shown in Table~\ref{tab:ds}, we evaluate the robustness of uncertainty estimation under distribution shift on ImageNet (CIFAR) variants. For methods that require training, we train the uncertainty estimators on the ImageNet-1K (CIFAR-10) training set and perform inference on the ImageNet-1K (CIFAR-10, CIFAR-100) validation set, ImageNet-R, and ImageNet-C (CIFAR-10-C, CIFAR-100-C) . Notably, distribution shift does not necessarily correspond to a monotonic decrease in zero-shot accuracy. For example, CLIP achieves even higher accuracy on ImageNet-R (74.3\%) than on the original ImageNet (66.7\%), while suffering a substantial degradation on ImageNet-C (46.8\%). This setting therefore provides a stringent test of whether UQ remains aligned with the model’s effective predictive competence, rather than merely reacting to the distribution shift itself.
We observe that parameterized uncertainty estimators that rely on fitting training-set patterns exhibit limited robustness under distribution shift. 
In particular, ViLU, which learns a static mapping from embeddings to risk scores using supervised failure signals, suffers from severe miscalibration once the target distribution deviates from the training distribution. 
For instance, while ViLU achieves strong performance on ImageNet (AUC 89.5), it degrades sharply on ImageNet-R and ImageNet-C, indicating that its learned decision boundaries fail to generalize beyond the training distribution. 
In contrast, post-hoc methods that directly operate on model outputs, such as MCM and Rel-U, exhibit more stable behavior across distributions, as they inherit the zero-shot generalization capability of the underlying VLM. 
However, their performance remains limited due to the lack of explicit modeling of the global feature-space geometry.
Our approach consistently achieves the best performance across all distributions. By dynamically estimating the embedding-space distribution and adapting GMM statistics to the target data stream, our framework avoids fitting training-specific uncertainty patterns. Instead, it derives uncertainty from the evolving distributional structure of the target distribution itself. As a result, both the Base and adaptation variants maintain high AUC and low FPR95 under all distribution shifts, demonstrating robust and reliable uncertainty estimation even when the data distribution changes substantially.

\begin{table}[h]
\centering
\caption{UQ Performance on image-label datasets with TTA-enhanced baselines.}
\label{tab:tta_analysis}
\resizebox{0.48\textwidth}{!}{%
\begin{tabular}{lcccccc}
\toprule
 & \multicolumn{2}{c}{ImageNet-1K} & \multicolumn{2}{c}{ImageNet-R} & \multicolumn{2}{c}{ImageNet-C} \\
\cmidrule(lr){2-3} \cmidrule(lr){4-5} \cmidrule(lr){6-7}
\textbf{Method} & \textbf{AUC$\uparrow$} & \textbf{FPR95$\downarrow$} 
& \textbf{AUC$\uparrow$} & \textbf{FPR95$\downarrow$} 
& \textbf{AUC$\uparrow$} & \textbf{FPR95$\downarrow$} \\
\midrule
ViLU  & 89.5±0.22 & 47.4±1.58 & 78.2±0.44 & 66.8±1.25 & 73.4±0.51 & 86.3±1.45 \\
MCM  & 80.8±0.35 & 71.3±1.05 & 78.3±0.42 & 65.2±1.12 & 72.1±0.48 & 81.6±1.18 \\
ViLU + DOTA \cite{dota} & 91.8±0.20 & 41.2±1.40 & 84.1±0.38 & 54.7±1.32 & 80.5±0.44 & 68.9±1.28 \\
MCM + DOTA & 83.5±0.33 & 66.8±1.15 & 81.9±0.41 & 60.3±1.21 & 78.6±0.47 & 74.2±1.35 \\
\rowcolor{lightblue} Ours (Base) & 91.2±0.18 & 43.8±1.45 & 88.9±0.24 & 48.6±1.22 & 84.7±0.31 & 65.1±1.28 \\
\rowcolor{lightpurple} Ours (Full) & \textbf{92.4±0.15} & \textbf{38.6±1.32} & \textbf{90.5±0.19} & \textbf{43.2±1.15} & \textbf{86.3±0.27} & \textbf{58.9±1.22} \\
\bottomrule
\end{tabular}%
}
\end{table}
\noindent\textbf{TTA-enhanced baselines.} As shown in Table~\ref{tab:tta_analysis}, to further investigate uncertainty estimation in the online setting, we design the following experiments. All methods are trained on ImageNet-1K and evaluated under distribution shifts from the perspective of test-time adaptation (TTA). We construct two composite baselines: ViLU + DOTA and MCM + DOTA, where DOTA is a test-time adaptation strategy. We evaluate all configurations on distribution-shift datasets including ImageNet-R and ImageNet-C. The experimental results demonstrate that our method consistently outperforms state-of-the-art uncertainty quantification methods and their combinations with test-time adaptation (e.g., ViLU+DOTA and MCM+DOTA) across all datasets. This indicates that uncertainty quantification directly derived from the data distribution achieves better performance than existing state-of-the-art ViLU combined with test-time adaptation, leading to more robust calibration.

\begin{table}[h]
\centering
\caption{UQ Performance on Multi-Label Datasets. The zero-shot mAP (in \%) of CLIP for each dataset is shown above the corresponding columns.}
\label{tab:multilabel_results}
\resizebox{0.48\textwidth}{!}{%
\begin{tabular}{lcccccc}
\toprule
 & \multicolumn{2}{c}{\textbf{VOC 2007}} 
 & \multicolumn{2}{c}{\textbf{MS-COCO}} 
 & \multicolumn{2}{c}{\textbf{NUS-WIDE}} \\
 & \multicolumn{2}{c}{78.4\%} 
 & \multicolumn{2}{c}{59.2\%} 
 & \multicolumn{2}{c}{46.7\%} \\
\cmidrule(lr){2-3} \cmidrule(lr){4-5} \cmidrule(lr){6-7}
\textbf{Method} 
& \textbf{AUC$\uparrow$} & \textbf{FPR95$\downarrow$} 
& \textbf{AUC$\uparrow$} & \textbf{FPR95$\downarrow$} 
& \textbf{AUC$\uparrow$} & \textbf{FPR95$\downarrow$} \\
\midrule
Doctor & 80.1±0.38 & 59.4±1.22 & 75.6±0.42 & 76.2±1.28 & 64.4±0.52 & 89.1±1.18 \\
Rel-U & 84.2±0.31 & 52.4±1.32 & 78.3±0.36 & 70.1±1.35 & 67.6±0.45 & 86.9±1.22 \\
BayesVLM & 83.9±0.32 & 51.7±1.35 & 79.8±0.34 & 67.5±1.38 & 74.6±0.38 & 73.9±1.45 \\
ViLU & 85.7±0.28 & 43.4±1.48 & 80.9±0.31 & 63.8±1.42 & 63.8±0.54 & 91.6±1.12 \\
\rowcolor{lightblue}
Ours (Base)  & 88.1±0.24 & 36.3±1.52 & 84.2±0.26 & 54.1±1.48 & 68.4±0.42 & 63.5±1.52 \\
\rowcolor{lightpurple}
Ours (Adaptation) & \textbf{89.4±0.21} & \textbf{34.9±1.55} & \textbf{85.6±0.22} & \textbf{52.7±1.51} & \textbf{71.2±0.38} & \textbf{61.8±1.58} \\
\bottomrule
\end{tabular}
}
\end{table}
\noindent\textbf{Complex Semantic Distributions.}
We evaluate our method on multi-label datasets to distinguish genuine uncertainty from the ambiguity caused by class co-existence. 
Crucially, we adopt a strict ``exact match'' criterion, where a prediction is correct only if all relevant labels are accurate. 
This renders Softmax-based baselines (e.g., MCM) ill-suited, as their inherent mutual-exclusivity assumption forces label competition, preventing accurate risk quantification for partial misclassifications.
As shown in Table~\ref{tab:multilabel_results}, our method achieves the best calibration across all datasets. 
The \textit{Adaptation} variant consistently improves upon the \textit{Base} baseline, validating the effectiveness of test-time distribution updates. 
Notably, on the semantically dense MS-COCO, our approach outperforms ViLU, demonstrating superior capability in isolating uncertainty from multi-label ambiguity. 
Even on the challenging NUS-WIDE dataset, where absolute performance naturally declines due to lower zero-shot accuracy, our method maintains a distinct relative advantage.

\begin{table}[h]
\centering
\caption{UQ performance on image-label datasets in OOD scenarios.}
\label{tab:ood_detection}
\resizebox{0.48\textwidth}{!}{%
\begin{tabular}{lcccccc}
\toprule
 & \multicolumn{2}{c}{\textbf{DTD}} & \multicolumn{2}{c}{\textbf{SVHN}} & \multicolumn{2}{c}{\textbf{Places365}} \\
\cmidrule(lr){2-3} \cmidrule(lr){4-5} \cmidrule(lr){6-7}
\textbf{Method} & \textbf{AUC}$\uparrow$ & \textbf{FPR95}$\downarrow$ & \textbf{AUC}$\uparrow$ & \textbf{FPR95}$\downarrow$ & \textbf{AUC}$\uparrow$ & \textbf{FPR95}$\downarrow$ \\
\midrule
MCM & 77.4 & 77.9 & 88.5 & 48.2 & 85.3 & 52.7 \\
KNN \cite{ood4}& 76.9 & 82.6 & 88.0 & 49.8 & 84.5 & 54.9 \\
GEN \cite{Gen}& 76.4 & 80.2 & 87.1 & 52.4 & 83.9 & 57.1 \\
DPU \cite{ood2}& 87.2 & 55.9 & 91.2 & 41.5 & 89.8 & 39.2 \\
Barrier \cite{OOD-Barrier} & 93.8 & 28.8 & 95.5 & \textbf{18.7} & 94.2 & 25.6 \\
\rowcolor{lightblue}
Ours(Base) & \textbf{95.1} & \textbf{24.3} & \textbf{96.7} & 22.1 & \textbf{96.1} & \textbf{21.4} \\
\bottomrule
\end{tabular}%
}
\end{table}
\noindent\textbf{Applied to OOD scenarios.}
Although UQ and OOD detection address distinct reliability objectives, our framework exhibits remarkable task generalization through the underlying GMM modeling. Specifically, without any architectural modifications for OOD tasks, our method naturally generalizes to OOD scenarios by leveraging the dual properties of GMM-derived evidence. The density-based evidence ($E_{\text{den}}$) inherently functions as a global sensor to capture the deviation of samples from the known manifold, effectively addressing OOD detection. Meanwhile, the ambiguity-based evidence ($E_{\text{amb}}$) provides fine-grained sensitivity to semantic confusion at decision boundaries within the established manifold, facilitating robust UQ. This seamless adaptation underscores that our framework captures intrinsic statistical structures rather than merely learning task-specific heuristics.
Table~\ref{tab:ood_detection} further confirms improved OOD detection with ImageNet-1K as in-distribution, where we outperform ViLU on DTD, SVHN, and Places365 (e.g., on DTD, $+1.3$ AUC and $-4.5$ FPR95). 
Overall, these results demonstrate that our framework improves both in-distribution failure prediction and robustness under distribution shifts, providing a practical Uncertainty quantification solution for real-world VLM deployment.

\subsection{Questions and Discussions}
\textbf{Q1: Does uncertainty depend on the model performance itself?}

\begin{figure}[!h]
    \centering
    \subfloat[Generalization FPR95]{
        \includegraphics[width=0.43\linewidth]{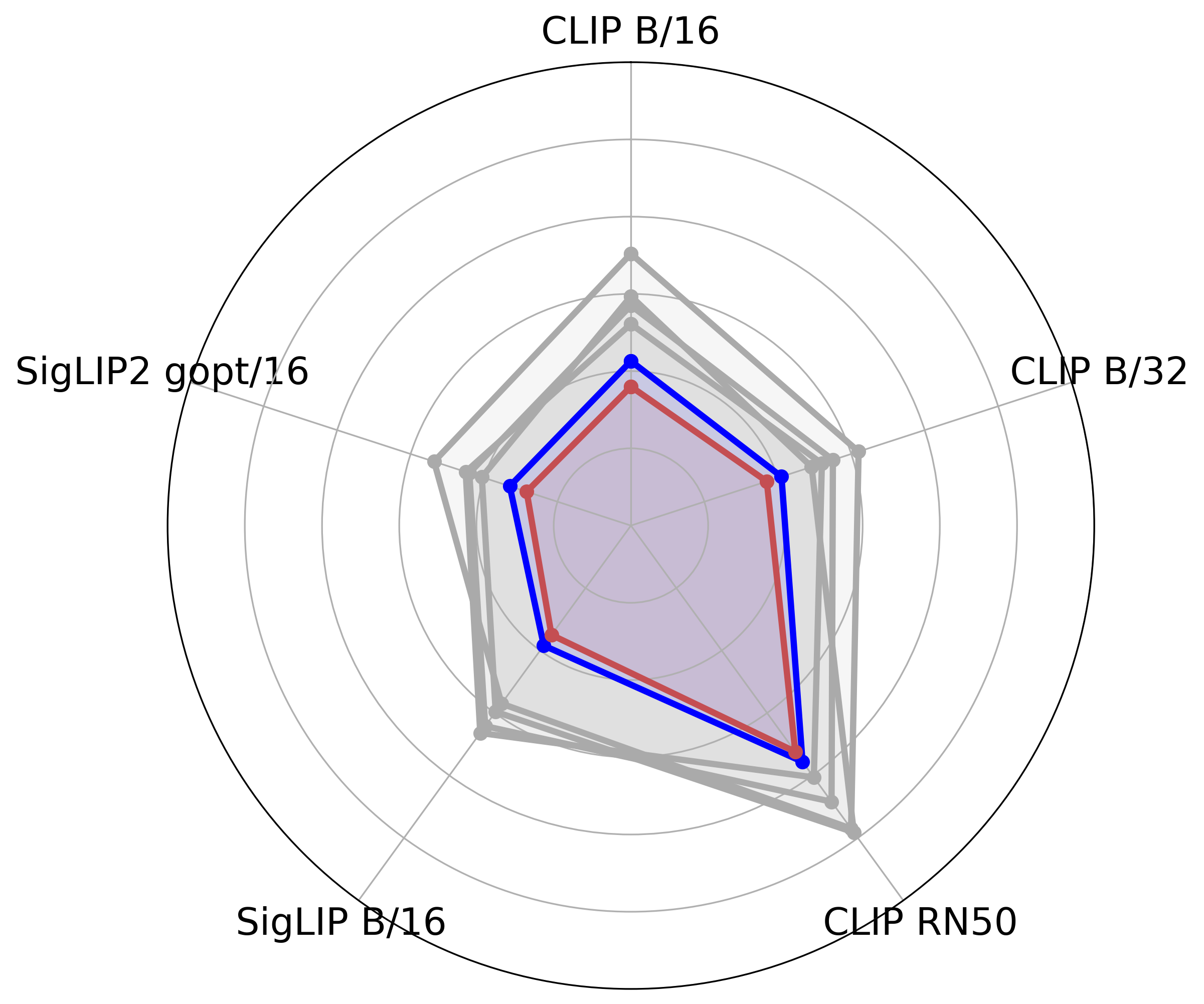}
        \label{subfig:radar_chart_fpr95}
    }
    \hfill
    \subfloat[Generalization AUC ]{
        \includegraphics[width=0.48\linewidth]{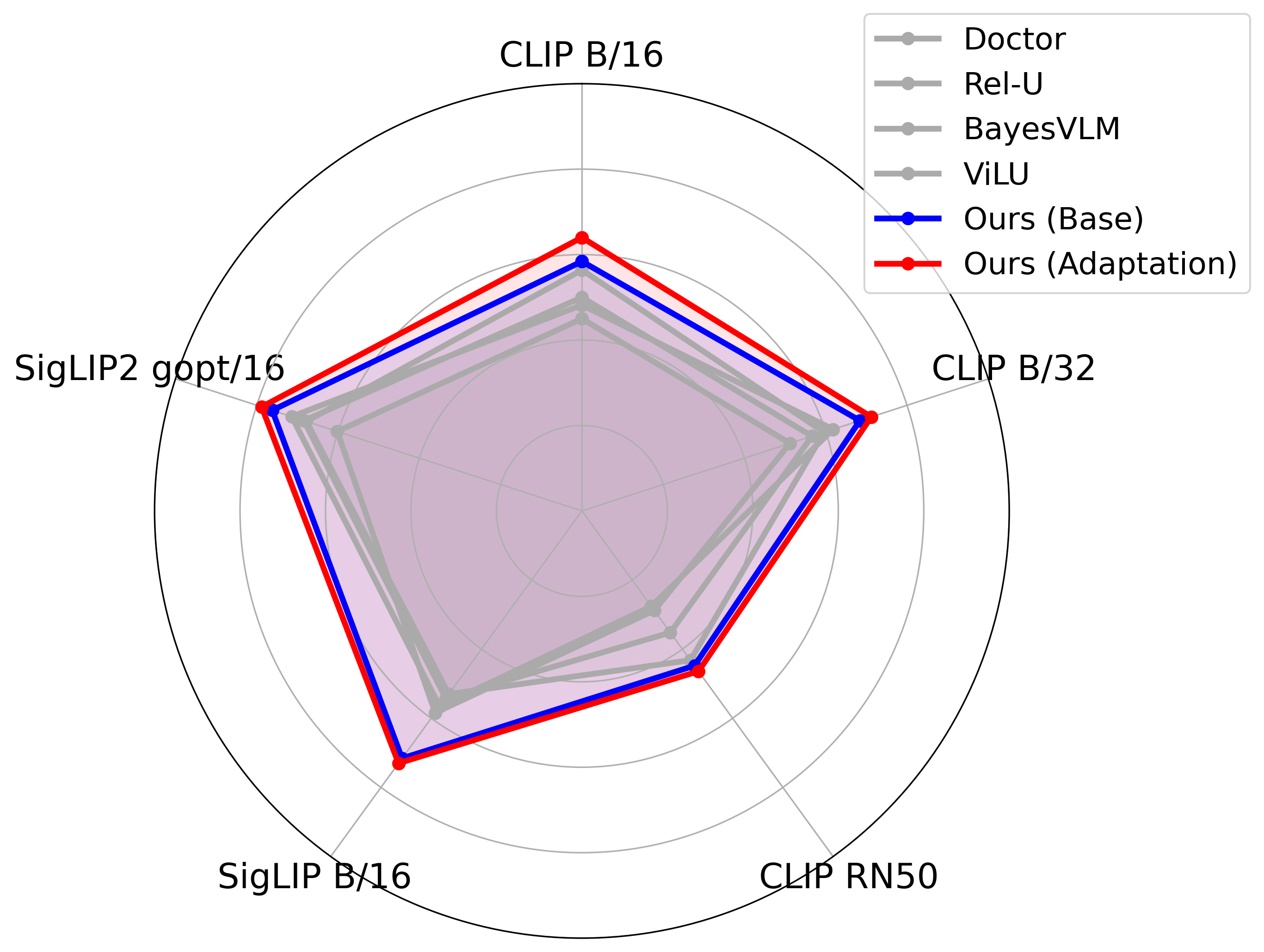}
        \label{subfig:radar_chart_auc}
    }
    \caption{Generalization performance of different methods across five vision backbones.}
    \label{fig:generalization}
\end{figure}

To some extent, this conclusion applies to all uncertainty quantification methods. A reliable backbone is a prerequisite for effective confidence estimation: no downstream uncertainty quantification module can correct the poor discriminative features generated by a weak backbone. However, as shown in Figure~\ref{fig:generalization}, our method exhibits excellent generalization performance across five vision backbones. Our two proposed methods achieve top results on nearly all backbones. Notably, Ours (Adaptation) achieves the highest AUC and lowest FPR95 across all five architectures, fully demonstrating its good robustness to different backbone capacities.

\textbf{Q2: Is the Gaussian assumption for high-dimensional CLIP embeddings reasonable?}

We fully agree that no strict multivariate Gaussian exists in high dimensions, and we do not claim CLIP embeddings strictly follow Gaussianity—we only use it as a reasonable approximation for class-conditional embeddings, supported by \cite{GDA1}. These works verify that CLIP embeddings present clear class-clustering on semantic manifolds, where Gaussian serves as an optimal and concise approximation. Our GMM only relies on first/second-order statistics (mean, covariance) to model distribution geometry, free from strict Gaussian constraints. Moreover, CLIP embeddings reside on low-dimensional semantic manifolds with effective dimensionality far lower than the apparent dimension, thus avoiding modeling failure caused by the curse of dimensionality.

\subsection{Ablation Study}
We examine the impact of each core component in Table~\ref{tab:ablation_refined}. Starting from a visual-only \textit{Baseline} (a), we observe that: (i) Impact of Distributional Context: Incorporating GMM-derived evidence (b) provides a substantial performance leap. This confirms that density and ambiguity signals offer critical context for identifying failure modes that raw visual features alone cannot capture. (ii) Conditioned Feature Fusion: Moving from simple concatenation to our \textit{adaptive modulation} (c) significantly enhances results. This demonstrates the superiority of conditioning the visual embeddings on distributional evidence, which enables the model to actively recalibrate visual activations based on their distributional validity rather than merely treating evidence as auxiliary input. (iii) Benefit of Dynamic Adaptation: The inclusion of Soft Risk Labels (d) and dynamic GMM Updates (e) yields the best performance. These components allow the model to learn fine-grained risk nuances and maintain alignment with the target distribution's evolving manifold, which is essential for robust uncertainty quantification.
\begin{table}[!h]
\centering
\caption{Ablation study of the proposed framework on Caltech101. We evaluate the contribution of distributional evidence (Evid.), adaptive modulation (Mod.), soft risk targets (Soft), and dynamic GMM updates (Update).}
\label{tab:ablation_refined}
\setlength{\tabcolsep}{8pt}
\resizebox{0.48\textwidth}{!}{%
\begin{tabular}{lcccc|cc}
\toprule
\textbf{Variant} & \textbf{Evid.} & \textbf{Mod.} & \textbf{Soft} & \textbf{Update} & \textbf{AUC$\uparrow$} & \textbf{FPR95$\downarrow$} \\
\midrule
(a) Baseline & - & - & - & - & 88.1 & 68.7 \\
(b) + Evidence & \checkmark & - & - & - & 92.8 & 24.2 \\
(c) + Modulation & \checkmark & \checkmark & - & - & 93.5 & 19.2 \\
(d) + Soft label & \checkmark & \checkmark & \checkmark & - & 94.7 & 15.8 \\
\rowcolor{lightblue}
(e) Full Model & \checkmark & \checkmark & \checkmark & \checkmark & \textbf{95.2} & \textbf{14.4} \\
\bottomrule
\end{tabular}
}
\end{table}

\subsection{Analysis}
\begin{figure}[!h]
    \centering
    \subfloat[ImageNet]{
        \includegraphics[width=0.9\linewidth]{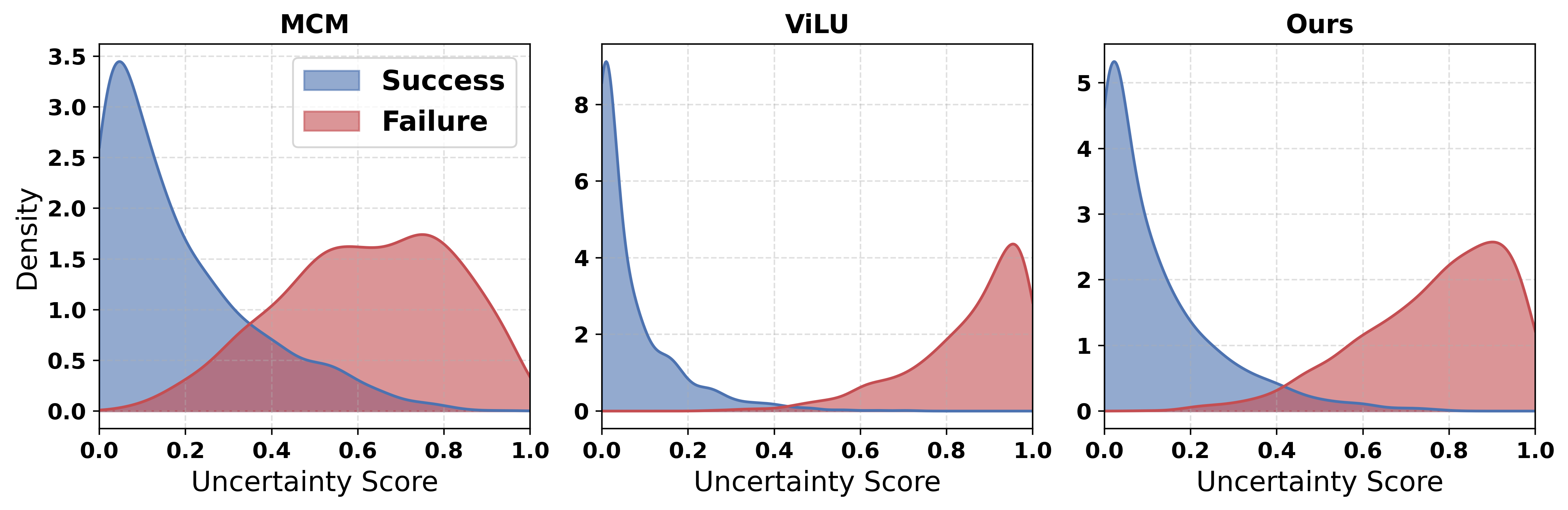}
        \label{subfig:tradition}
    }
    \hfill
    \subfloat[ImageNet-R (Distribution shift)]{
        \includegraphics[width=0.9\linewidth]{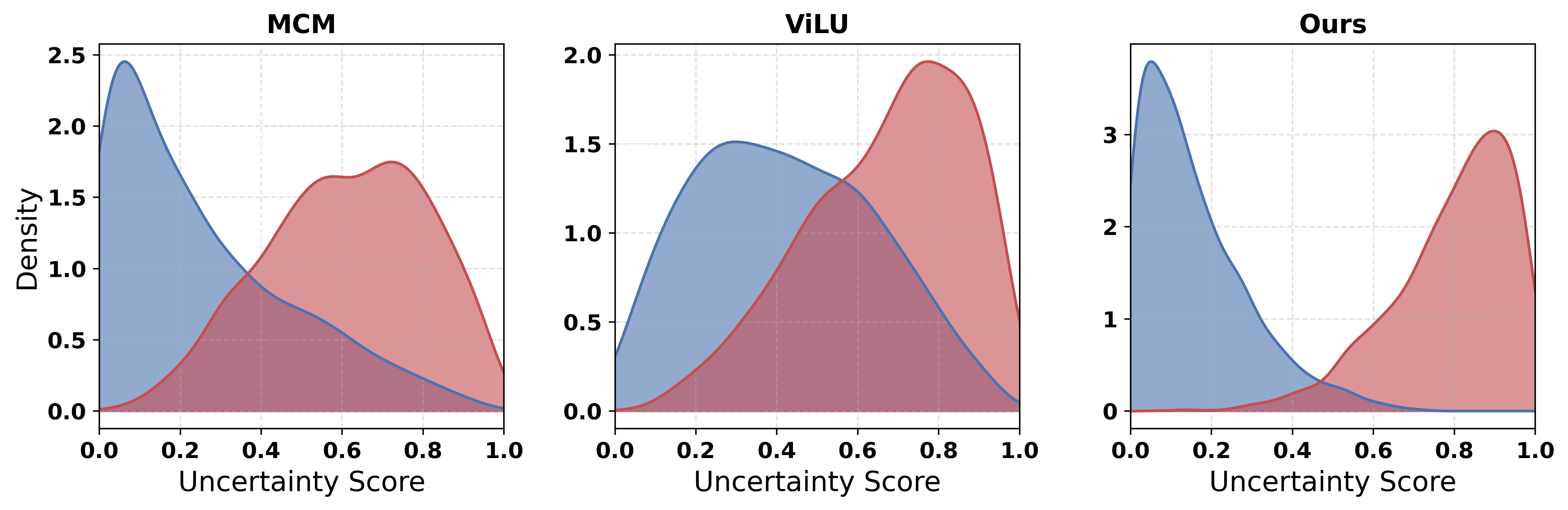}
        \label{subfig:domain_adaptation}
    }
    \caption{Uncertainty Score Distributions on ImageNet and ImageNet-R.} 
    \label{fig:Uncertainty_distribution} 
\end{figure}

\begin{figure}[!h]
    \centering
    \subfloat[Training data needed]{
        \includegraphics[width=0.46\linewidth]{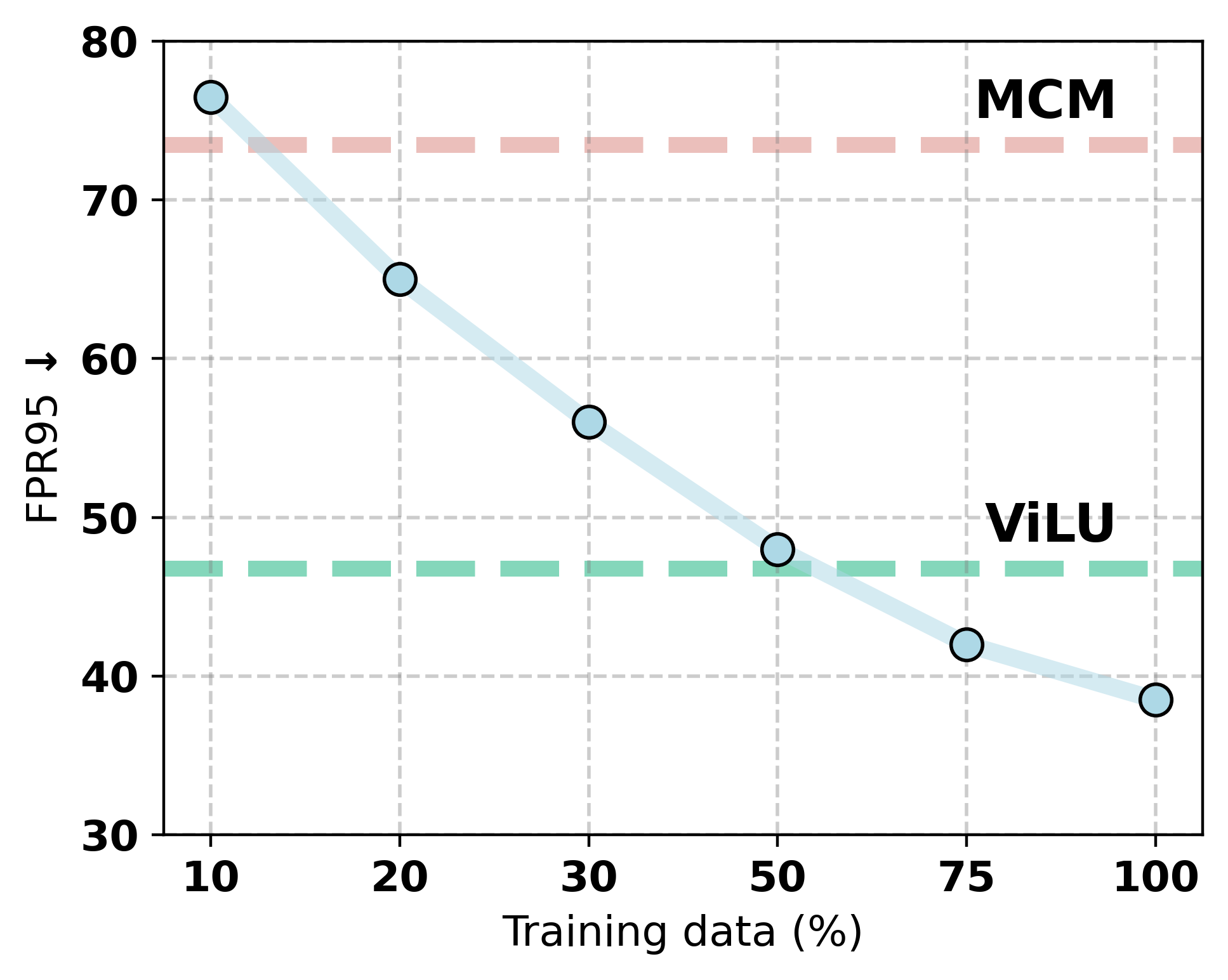}
        \label{subfig:}
    }
    \hfill
    \subfloat[Infer data batch]{
        \includegraphics[width=0.46\linewidth]{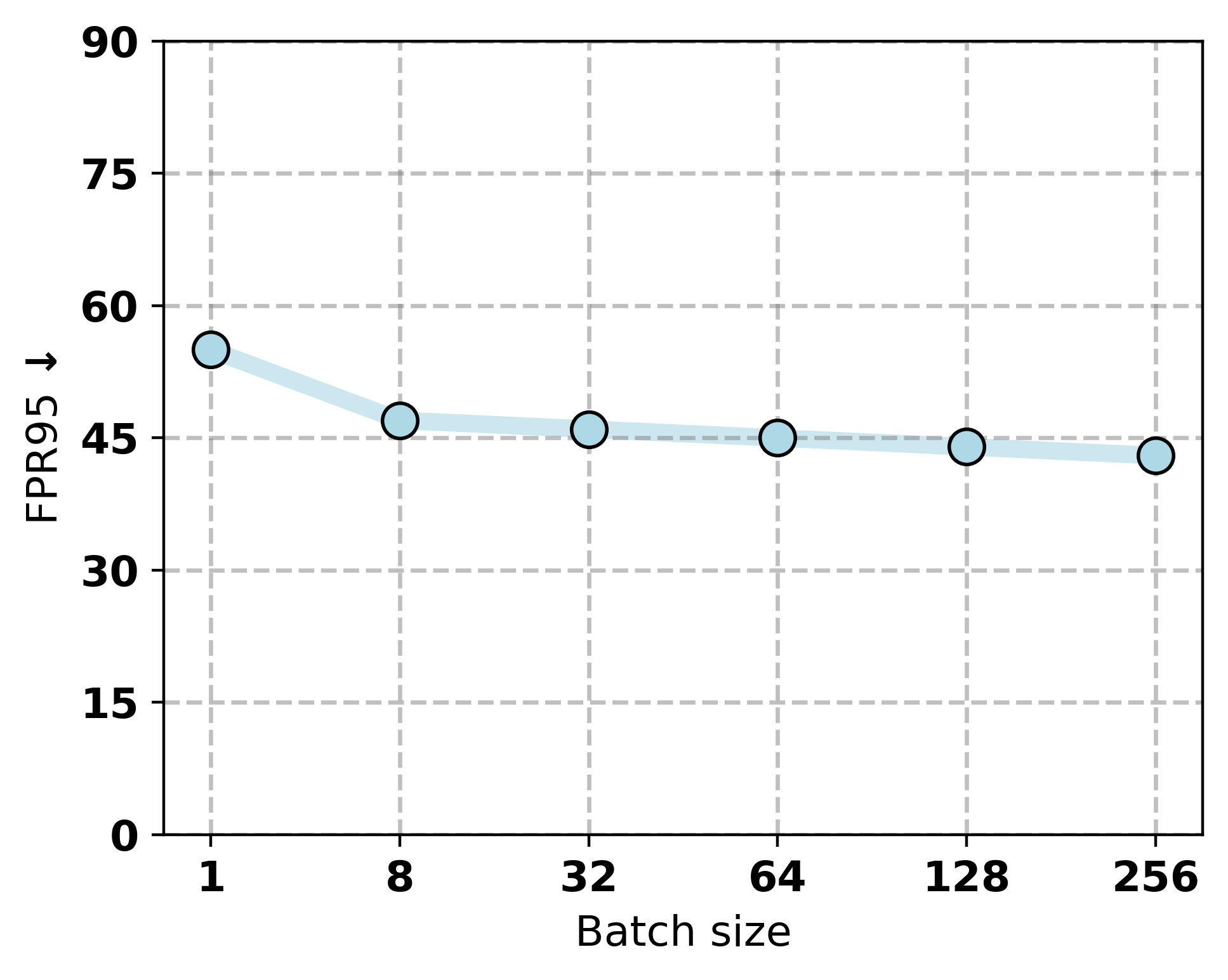}
        \label{subfig:}
    }
    \caption{Efficiency and Robustness Analysis.}
\label{fig:efficiency}
    \label{fig:data_efficiency} 
\end{figure}

Figure~\ref{fig:Uncertainty_distribution} highlights the vulnerability of ViLU under distribution shift: its dependence on static semantic embeddings leads to a misalignment between the frozen text anchors and the shifted visual features. Consequently, it fails to distinguish between true semantic novelty and mere stylistic variations (e.g., in ImageNet-R), causing miscalibrated risk scores. Conversely, our framework shifts the paradigm from static matching to dynamic adaptation. By modeling the visual manifold via a dynamic GMM and explicitly accounting for semantic ambiguity, we enable the uncertainty estimation to actively synchronize with the target distribution, ensuring robust calibration even in unsupervised environments.
As illustrated in Figure~\ref{fig:data_efficiency}, our framework demonstrates remarkable efficiency in both training and inference phases.
First, regarding training efficiency (Figure~\ref{fig:data_efficiency}a), our method rapidly converges, surpassing the strong zero-shot baseline (MCM) using as little as 20\% of the training data. We attribute this to our statistical GMM modeling: unlike discriminative methods (e.g., ViLU) that require abundant samples to delineate complex boundaries, estimating sufficient statistics (means and covariances) is inherently sample-efficient. Furthermore, the semantic-aware soft target accelerates convergence by providing dense, continuous gradient signals from each sample.
Second, regarding inference stability (Figure~\ref{fig:data_efficiency}b), our method maintains robust performance across varying batch sizes. Notably, even at a batch size of 1 (online streaming setting), the model achieves competitive FPR95, which further improves and saturates around a batch size of 32. This confirms that our EMA update mechanism is stable and does not strictly require large buffers, making it highly suitable for real-time, resource-constrained deployment.

Additional qualitative and parameter analyses are provided in Appendix A.4. 
The case study under distribution shift shows that, although CLIP can remain robust or even improve its recognition performance on shifted data, existing uncertainty estimation methods such as ViLU may still suffer substantial degradation due to their reliance on static source-domain representations. 
In contrast, our distribution-aware framework consistently yields more reliable uncertainty estimates by adapting to the evolving representation structure of the target distribution. 
We further analyze the sensitivity to the number of GMM components $K$ and observe that performance improves as $K$ increases and becomes stable when $K$ approaches the number of classes $C$. 
Accordingly, we set $K=C$ as the default configuration in all experiments.
\section{Conclusion}
This work introduces the Dynamic Distribution-Aware Uncertainty Quantification framework (DDA-UQ) for Vision-Language Models, which overcomes the limitations of static uncertainty estimation. By leveraging a Gaussian Mixture Model to capture the embedding space structure, DDA-UQ shifts uncertainty quantification from a fixed mapping to a dynamic, distribution-aware process. Extensive experiments validate that our framework significantly improves robustness to distribution shifts, providing a reliable safety mechanism for deploying VLMs in real-world, non-stationary environments.

\section*{Acknowledgements}
We would like to thank the anonymous reviewers for their insightful comments. 
This work is supported by the JiangSu Natural Science Foundation under Grant No.BK20251989; the National Natural Science Foundation of
China under Grants Nos. 62172208, 62441225, 61972192; the Fundamental and Interdisciplinary Disciplines Breakthrough Plan of the Ministry of Education of China (No.JYB2025XDXM118); the “111 Center” (No. B26023). 
This work is partially supported by Collaborative Innovation Center of Novel Software Technology and Industrialization.

\bibliographystyle{ACM-Reference-Format}
\bibliography{sample-base}

\newpage

\end{document}